\documentclass[letterpaper, 10 pt, conference]{ieeeconf}  % Comment this line out if you need a4paper

\IEEEoverridecommandlockouts                              % This command is only needed if 
\usepackage{times}
\usepackage{epsfig}
\usepackage{graphicx}
\usepackage{amsmath}
\usepackage{amssymb}
\usepackage{arydshln}
\usepackage{siunitx}
\usepackage{multirow,array}
\usepackage{pifont}
\usepackage{xcolor}

\usepackage[utf8x]{inputenc}
\usepackage{pgfplots}
\usepackage{pgfplotstable}
\pgfplotsset{compat=newest}
\usepackage{tikz}
\usetikzlibrary{backgrounds}
\usepackage{color}
\usepackage{booktabs,siunitx} % better tables
\usepackage{makecell}
\usepackage[caption=false]{subfig}
\usepackage[noadjust]{cite}
\usepackage{array,multirow,graphicx}
\usepackage{rotating}

\usepackage{hyperref}

\definecolor{30}                    {RGB}{188, 193, 245}
\definecolor{50}                    {RGB}{144, 152, 238}
\definecolor{80}                    {RGB}{100,111,232}
\definecolor{120}                   {RGB}{55,70,225}

\definecolor{Light Near}            {RGB}{255,198,128}
\definecolor{Light Far}             {RGB}{255,152,26}

\definecolor{Crossing Near}         {RGB}{153,230,153}
\definecolor{Crossing Far}          {RGB}{71,210,71}

\definecolor{Left}                  {RGB}{255,128,128}
\definecolor{Straight}              {RGB}{255,77,77}
\definecolor{Right}                 {RGB}{255,26,26}

\definecolor{Enter}                 {RGB}{219, 190, 244}
\definecolor{Inside}                {RGB}{195,146, 236}
\definecolor{Exit}                  {RGB}{170,103, 229}
\definecolor{None}                  {RGB}{146,60, 221}
\newcommand{\YES}{\ding{51}}

\newcommand*\rot[1]{\rotatebox{90}{#1}}

\begin{document}

%%%%%%%%% TITLE
\title{Learning Accurate and Human-Like Driving using Semantic Maps and Attention}

\author{Simon Hecker$^{1}$\quad  Dengxin~Dai$^{1}$ \quad Alexander Liniger$^{1}$ \quad  Luc~Van~Gool$^{1,2}$  % <-this % stops a space
\\
$^1$ CVL, ETH Zurich \hspace{5mm}
$^2$ESAT-PSI, KU Leuven \\
\{heckers,alex.liniger,dai,vangool\}@vision.ee.ethz.ch
\\
\thanks{*The work is supported by Toyota via the project TRACE-Zurich.}
}
% \author{Simon Hecker\\
% Computer Vision Lab\\
% ETH Z\"urich\\
% {\tt\small heckers@vision.ee.ethz.ch}
% \and
% Dengxin Dai\\
% Computer Vision Lab\\
% ETH Z\"urich\\
% {\tt\small dai@vision.ee.ethz.ch}
% \and
% Luc Van Gool\\
% Computer Vision Lab\\
% ETH Z\"urich\\
% {\tt\small vangool@vision.ee.ethz.ch}
% }

\maketitle
%\thispagestyle{empty}

%%%%%%%%% ABSTRACT
\begin{abstract}
This paper investigates how end-to-end driving models can be improved to drive more accurately and human-like. To tackle the first issue we exploit semantic and visual maps from HERE Technologies and augment the existing Drive360 dataset with such. The maps are used in an attention mechanism that promotes segmentation confidence masks, thus focusing the network on semantic classes in the image that are important for the current driving situation. Human-like driving is achieved using adversarial learning, by not only minimizing the imitation loss with respect to the human driver but by further defining a discriminator, that forces the driving model to produce action sequences that are human-like. Our models are trained and evaluated on the Drive360 + HERE dataset, which features $60$ hours and $3000$ km of real-world driving data. Extensive experiments show that our driving models are more accurate and behave more human-like than previous methods. %The resources of this work will be released on the \href{http://people.ee.ethz.ch/~tracezuerich/learning-to-drive/human-like/}{project page}

% Autonomous vehicles are more likely to be accepted if they drive accurately, comfortably, but also similar to how human drivers would. This is especially true when autonomous and human-driven vehicles need to share the same road. The main research focus thus far, however, is still on improving driving accuracy only. This paper formalizes the three concerns with the aim of accurate, comfortable and human-like driving. Three contributions are made in this paper. First, numerical map data from HERE Technologies are employed for more accurate driving; a set of map features -- which are believed to be relevant to driving -- are engineered to navigate better. Second, the learning procedure is improved from a pointwise prediction to a sequence-based prediction and passengers' comfort measures are embedded into the learning algorithm. Finally, we take advantage of the advances in adversary learning to learn human-like driving; specifically, the standard L1 or L2 loss is augmented by an adversary loss which is based on a discriminator trained to distinguish between human driving and machine driving. Our model is trained and evaluated on the Drive360 dataset, which features $60$ hours and $3000$ km of real-world driving data. Extensive experiments show that our driving model is more accurate, more comfortable and behaves more like a human driver than previous methods. The resources of this work will be released on the \href{http://people.ee.ethz.ch/~tracezuerich/learning-to-drive/human-like/}{project page}. 

 %our proposed model and actual human driver behaviors are highly similar with respect to gap acceptance in intersections.
\end{abstract}

%%%%%%%%% BODY TEXT
\section{Introduction}
\label{sec:intro}
%-------------------------------------------------------------------------

Over the last few decades autonomous driving has seen dramatic advances, from the humble beginnings \cite{dickmanns1987autonomous}, over the DARPA challenges \cite{darpa:grand,darpa:urban}, to today's autonomous driving companies which have driven tens of millions of miles autonomously on public roads. 

These massive gains were achieved by improving all the components of an autonomous car over the years. Advances were not limited to the hardware, but also to the algorithms necessary to drive a car. Normally these algorithms are large software stacks that are built using multiple layers, such as perception, localization, motion planning, and control, see \cite{darpa:urban,AD:Boss:08}. However, due to the complexity of such stacked systems, in recent years we have seen a rise of end-to-end driving models that solve autonomous driving. These driving models are an elegant alternative by directly mapping sensor inputs to driving actions \cite{nvidia:driving:16,end:driving:imitation:18,drive:surroundview:route:planner}.

Most works on end-to-end driving models use simplistic sensor setups, when compared to traditional autonomous driving stacks \cite{darpa:urban}. However, recent work showed, that rendered maps can improve the performance of end-to-end driving models \cite{drive:surroundview:route:planner,end-to-end:map:variational}, and if HD-maps are available that they can be even used as a fundamental part of the end-to-end driving model \cite{chauffeurnet:18,end-to-end:uber:motionplanner}. 

End-to-end driving models can be deployed to either maneuver an autonomous car, act as a sanity checker of a traditional stack in a tandem approach or be used to evaluate human driving in mobility as a service applications (s.a. taxi driver fatigue). But as such, they not only need to be able to drive \emph{accurately}, but should also drive \emph{human-like}, as this is believed to increase the acceptance of autonomous cars~\cite{human:like:driving:simulator:01,realistic:driving:03,human:like:motion:planning:17} and improve human driver evaluation capability.

In this work, we tackle both accurate driving, using high fidelity semantic maps, and human-like driving. This also directly defines our three main contributions: First, the Drive360 dataset introduced in \cite{drive:surroundview:route:planner} is extended with high precision semantic maps from HERE Technologies. To the best of our knowledge, this is the first large scale dataset, suited for training end-to-end driving models that includes high precision semantic maps. Second, we propose a novel way to include these semantic maps in the end-to-end driving model using an attention mechanism that can promote different confidence masks of a semantic segmentation network, allowing to combine the map information with semantic information in the image. Third, to achieve human-like driving we propose to use adversary learning to teach the car about human driving styles. Specifically, a discriminator is trained, together with our driving model, to distinguish between human driving and our ``machine" driving. This allows us to train for accurate and human-like driving at the same time. A preliminary version of this work has been released on arXiv before with substantial differences \cite{driving:comfortable}.

\section{Related Work}

%Our work is related to 1) learning driving models, 2) navigation maps, 3) ride comfort, and 4) human-like driving. 
%We present the related work in four groups. 
\noindent
\textbf{Driving Models}.
Significant progress has been made in autonomous driving in the last few years. Classical approaches require the recognition of all driving-relevant objects, such as lanes, traffic signs, traffic lights, cars and pedestrians, and then perform motion planning, which is further used for final vehicle control~\cite{AD:Boss:08}. These type of systems are sophisticated, and represent the current state-of-the-art for autonomous driving, however they are hard to maintain and prone to error accumulation along the pipeline. 
%Most systems also need to use diverse sensors, such as cameras, laser scanners, radar, GPS and high-definition maps~\cite{autonomous:vehicle:guide:policymakers}. 

End-to-end driving methods, on the other hand, construct a direct mapping from the sensory input to the actions. The idea can be traced back to the 1980s~\cite{network:autonomous:1980}. Other more recent end-to-end examples include~\cite{LeCun:driving:05,nvidia:driving:16,lidar:end:driving:18,end:driving:eventcamera:18,end:driving:imitation:18,drive:surroundview:route:planner,E2E:auxiliary:aaai18}. In \cite{nvidia:driving:16}, the authors trained an end-to-end method with a collection of front-facing videos. The idea was extended later on by using a larger video dataset~\cite{end:driving:16}, by adding side tasks to regularize the training~\cite{end:driving:16,E2E:auxiliary:aaai18}, by introducing directional commands~\cite{end:driving:imitation:18} and route planners~\cite{drive:surroundview:route:planner} to indicate the destination, by using multiple surround-view cameras to extend the visual field~\cite{drive:surroundview:route:planner}, by adding synthesized off-the-road scenarios~\cite{chauffeurnet:18}, and by adding modules to predict when the model fails~\cite{driving:failure:prediction}. The main contributions of this work, namely using semantic map data, either directly or through an attention mechanism, and rendering human-like driving in an end-to-end learning framework, are complementary to all methods developed before.   

There are also methods dedicated to robust transfer of driving policies from a synthetic domain to the real world domain~\cite{driving:policy:transfer:18,drive:sim2real:19}. Some other works study how to better evaluate the learned driving models~\cite{challenges:AV:testing:16,offline:evaluation:driving:18}. Those works are complementary to our work. Other contributions have chosen the middle ground between traditional pipe-lined methods and the monolithic end-to-end approach. They learn driving models from compact intermediate representations called affordance indicators such as \emph{distance to the front car} and \emph{existence of a traffic light}~\cite{deep:driving,conditional:affordance:learning:18}. Our engineered features from semantic maps can be considered as some sort of affordance indicators. Recently, reinforcement learning for driving~\cite{reinforcement:learning:driving,DRL:dirivng:17,drive:in:a:day:19} and learning to drive in simulators \cite{Codevilla_2019_ICCV,Ohn-Bar_2020_CVPR} have both received increased attention. %The trend is especially fueled by the release of multiple driving simulators~\cite{AirSim:17,CARLA:simulator}.  
%Methods have also been developed to explain how the end-to-end networks work for the driving task~\cite{explaining:end:driving:17} and to predict when they fail~\cite{driving:failure:prediction}.

\textbf{Navigation Maps}. Increasing the accuracy and robustness of self-localization on a map~\cite{HD:map:10,HD:map:12,traffic:rules:AV:15} and computing the fastest, most fuel-efficient trajectory from one point to another through a road network~\cite{Driving:knowledge:world:11,route:planning,scenic:driving:route:13} have been popular research fields for many years. By now, navigation systems are widely used to aid human drivers or pedestrians. Yet, their integration for learning driving models has not received a lot of attention in the academic community, mainly due to limited accessibility~\cite{drive:surroundview:route:planner}.  
We integrate industrial standard semantic maps -- from HERE Technologies -- into the learning of our driving models. We show the advantage of using these maps either in a straightforward late fusion approach or via a map-based attention module. Similar map features have been used recently in an ADAS system for motorcycle \cite{itsc19_curve_guardian_motorcycle}.

% 
%A preliminary version of this work has been released on arXiv \cite{driving:comfortable}.  

\textbf{Attention}. In recent years several researchers propose to use different attention mechanism, for end-to-end driving models \cite{attention:visual:Kim_2017_ICCV,attention:objectcentric:ad:icra19,attention:advice:Kim_2019_CVPR}. In \cite{attention:visual:Kim_2017_ICCV} visual attention map is used to visualize the focus of the network. In \cite{attention:objectcentric:ad:icra19}, the attention is more guided and can only promote detected objects. Whereas the former is vision-based, in \cite{attention:advice:Kim_2019_CVPR}, natural language based advise to the network is used to focus the network's attention. Our approach differs in the sense that it does not use visual or language-based attention guidance but instead utilizes the rich information present in semantic maps to promote visual object classes based on the driving location.  

\textbf{Human-Like Driving}. A large body of work has studied human driving styles~\cite{vehicle:corpora:driver:behavior:11,driving:style:survey:15}. Also statistical approaches were employed to evaluate human drivers and to suggest improvements~\cite{driving:behavior:smartphones:12,driving:behavior:evaluation:13}. Some work has even studied human-like motion planning of autonomous cars, but it was constrained to simulated scenarios~\cite{human:like:driving:simulator:01,realistic:driving:03}. In \cite{driving:styles:AV:15} a cost function that can generate human-like driving was learned using inverse reinforcement learning.  Instead of learning a cost, in our work we rely on adversarial learning to force our driving model to generate action sequences that come from the same distribution as human action sequences. Note that using adversarial learning is not a new concept in imitation learning \cite{GAIL}. However, using a discriminator to force the policy to learn human like action sequences is new and compared to \cite{GAIL} our approach is applicable to systems where environment interactions are restricted.

\setlength{\textfloatsep}{10pt}
\begin{table*}[tb]
\footnotesize
\vspace{1.3mm}
\begin{tabular*}{\linewidth}{l @{\extracolsep{\fill}}ll} \toprule
Group and Name & Range  & Description \\ \midrule
1.a \textit{distanceToIntersection} & $[0m, 250m]$ & Road-distance to next intersection encounter.  \\
1.b \textit{distanceToTrafficLight} & $[0m, 250m]$ & Road-distance to next traffic light encounter. \\ 
1.c \textit{distanceToPedestrianCrossing} & $[0m, 250m]$ & Road-distance to next pedestrian crossing encounter. \\ 
1.d \textit{distanceToYieldSign} & $[0m, 250m]$ & Road-distance to next yield sign encounter. \\ 
2.a \textit{speedLimit} & $[0km/h, 120km/h]$ & Legal speed limit for road sector.\\
2.b \textit{freeFlowSpeed} & $[0 km/h, \infty km/h)$ & Average driving speed based on underlying road geometry. \\ 
3.a \textit{curvature} & $[0 m^{-1}, \infty m^{-1})$ & Inverse radius of the approximated road geometry by means\\ 
4.a \textit{turnNumber} & $[0, \infty)$ & Index of road at next intersection to travel (counter-clockwise).   \\ 
5.a \textit{ourRoadHeading} & $[\ang{-180}, \ang{180})$ &  Relative heading of road that car will take at next intersection.\\
5.b \textit{otherRoadsHeading} & $(\ang{-180}, \ang{180})$ &  Relative heading of all other roads at next intersection.\\ 
6.a - 6.e \textit{futureHeadingXm}  & $[\ang{-180}, \ang{180})$ & Relative heading of map matched GPS coordinate in  $X\in\{1,5,10,20,50\}$ meters. \\
\bottomrule
% $X \in \{1, 5, 10, 20, 50\}$ & $[\ang{-180}, \ang{180})$ &  Relative heading of map matched GPS coordinate in 5 meters.\\
% 6.c \textit{futureHeading10m} & $[\ang{-180}, \ang{180})$ & Relative heading of map matched GPS coordinate in 10 meters.\\
% 6.d \textit{futureHeading20m} & $[\ang{-180}, \ang{180})$ & Relative heading of map matched GPS coordinate in 20 meters. \\
% 6.e \textit{futureHeading50m} & $[\ang{-180}, \ang{180})$ & Relative heading of map matched GPS coordinate in 50 meters.\\
\end{tabular*}
\caption{A summary of HERE map data used in this work. 
}
% Speed at which vehicles can travel current road sector when \\ there is no traffic congestion. Gives 
% \\ of circles at road geometry points.
% Intersection is defined as a branch in the map tree.
% \\ i.e. if $2$, take the $2^{nd}$ road from left.
%%%% Corresponding color values to image
% 5b: ED2024
% 5a: B11F24
% 6d: 314D78
% 6a: 4B6EB5
% 2a: F8971D
% 2b: F37C21
% 1c: 95C93D
% 1d: 519A43
\label{table:hereFeatures}
\vspace{-10mm}
\end{table*}

\section{Approach}
In this section, we present our contributions: we extend the existing Drive360 dataset with semantic and visual maps from HERE, we improve state-of-the-art end-to-end driving models with maps and attention modules and finally we formulate a new adversarial training strategy to enable human-like driving. 

\subsection{Drive360 + HERE}
\label{sec:dataset}
In this section, we describe how we augment real-world driving data, specifically the Drive360 dataset~\cite{drive:surroundview:route:planner} with additional map data from HERE Technologies. All our  HERE map data will be made publicly available.

\subsubsection{Obtaining HERE Map Data}
The original Drive360 dataset features $60$ hours of real-world driving data over $3000$ km. It offers recordings of eight roof-mounted cameras, a rendered TomTom visual-route planning module and vehicle speed and steering wheel angle recorded from the human driver. However, this dataset comes with two significant shortcomings, 1) the visual-route planner rendered by a cell phone app is not always synchronized precisely to the camera streams and 2) it lacks precise semantic map information. 

To tackle these two issues, we augment the Drive360 dataset with HERE Technologies map data to supply an accurately synchronized visual-route planning module and additional semantic map information. 
Drive360 offers a time stamped GPS trace for each route recorded. We use a path-matcher based on a hidden markov model employing the Viterbi algorithm \cite{forney1973viterbi} to calculate the most likely path traveled by the vehicle during dataset recording, thus, snapping the recorded GPS trace to the underlying road network. This improves our vehicle localization accuracy significantly over the recorded GPS trace, especially in urban environments where the raw GPS signal may be weak and noisy. Through the path-matcher we obtain a map matched GPS coordinate for each time stamp, which is then used to query the HERE Technologies map database to obtain the various types of map data. As we are augmenting an existing dataset, using a path-matcher is our best option for obtaining precise localization given the existing data.

\begin{figure}
\centering
\vspace{1.5mm}
\includegraphics[width=0.75\linewidth]{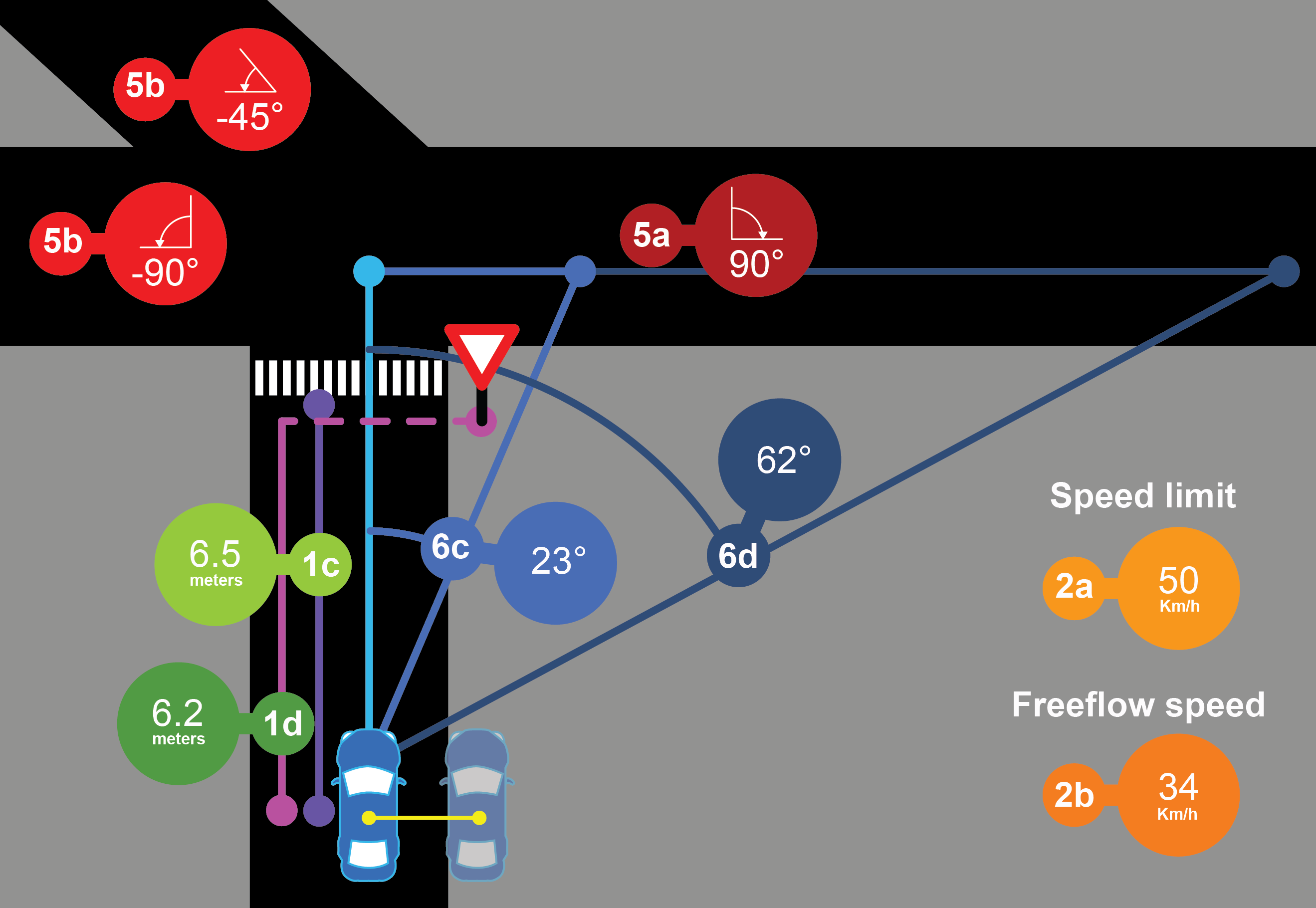}
\caption{An illustration of HERE map features used in this work. Please refer to Table~\ref{table:hereFeatures} for a detailed feature description.} 
\label{fig:here_features}
%%%% Corresponding color values to image
% 5b: ED2024
% 5a: B11F24
% 6d: 314D78
% 6a: 4B6EB5
% 2a: F8971D
% 2b: F37C21
% 1c: 95C93D
% 1d: 519A43
\vspace{-1mm}
\end{figure}

Note that the particular HERE map we query is not considered a true HD-map as used in today's Geo-fenced autonomous driving applications. However, in contrast, the employed map is available for large parts of the world and has significantly more coverage than the geographically restricted HD-maps. In fact, the employed HERE map has coverage for all Drive360 routes including those in rural parts of Switzerland, where, in the foreseeable future, corresponding HD-maps will not be available. This enables the development of autonomous driving in areas with no HD-map coverage.

\subsubsection{Visual Rendering} 
\label{sec:dataset:visual_render}
Employing the map matched GPS coordinates we can render a new visual-route planning module using HERE maps that is better temporally aligned to the rest of the Drive360 data streams. We render a fixed size map segment, with a real-world size of 200m by 200m, where the current map-matched position is located in the center of the image (575px by 575px), and the map is oriented with respect to the cars heading. The rendered map contains a detailed rendering of the road network with the past and future route highlighted.

\subsubsection{Semantic Maps} 
\label{sec:dataset:semantic_maps}
HERE Technologies has generated an abundant amount of semantic map data. We selected $15$ types of data of $6$ categories, as described in Table \ref{table:hereFeatures}. All features belonging to category 1 will be capped at 250m, for example no $distanceToTrafficLight$ feature is given if the next traffic light on route is further than 250m from the current map matched position. The features of category 5 specify the relative heading of all roads exiting the next approaching intersection, with regard to the map matched entry heading, see Figure~\ref{fig:here_features}. The features of category 6 specify the relative heading of the planned route a certain distance in advance. This relative heading is only calculated with map matched positions. The relative heading is dependent on the road geometry and the route taken, see Figure~\ref{fig:here_features}. Using more types of map data constitutes our future work. 

\subsubsection{Towards Performance Interpret-ability} An extremely useful feature of the HERE semantic map is that they enable driving-scenario specific model training and evaluation by defining data subsets. The original Drive360 dataset came without driving-scenario annotation. Therefore, it was infeasible to select a subset of, for example, traffic light or pedestrian crosswalk samples. However, with HERE data annotation it is now possible to filter the Drive360 dataset along any of the given features, providing the community with much finer control over the training or evaluation set samples. For example, combining the two filters \textit{distance to traffic light less than 50m} and \textit{vehicle stopped}, one, with high probability, obtains mainly instances where the vehicle is stopped at a red light. A subsequent evaluation on this subset allows for better performance interpret-ability of the driving model in selected driving scenarios.

\subsection{Driving Models}
\label{sec:approach:driving_models}
The community has developed promising driving models based on camera data~\cite{end:driving:16,end:driving:imitation:18,drive:surroundview:route:planner}. However, the focus has mainly been on perception, not so much navigation. Thus far, the representations for navigation are either primitive directional commands in a simulation environment~\cite{end:driving:imitation:18,conditional:affordance:learning:18} or rendered top down views of planned routes in real-world environments~\cite{drive:surroundview:route:planner, end-to-end:uber:motionplanner, chauffeurnet:18, end-to-end:map:variational}. 

\subsubsection{Basic Driving Model}
% End-to-End driving models in their basic form, employ deep neural networks to predict the control inputs for a car directly from camera data \cite{}. Even though such models have been shown to be able to learn to drive a car, recent work showed that adding navigation information can improve the driving model \cite{}. 
Our basic driving model is adopted from the approach presented in~\cite{drive:surroundview:route:planner} and combines front-facing camera data with rendered navigation information. The model takes a sequence of past images and map renders and predicts the steering angle $\delta$ and velocity $v$ for a future time step.

More precisely, we use the following notation in our end-to-end driving model, let $I_t$ denote the $t$-th frame in the front-facing video stream, $M_t$ the $t$-th frame of the rendered navigation maps. We regard these images as our observation and group them as $O_t = (I_t,M_t)$. As mentioned our model should compute the steering angle $\delta$ and velocity $v$, thus our actions at the $t$-th time step, are given by $a_t = (\delta_t,v_t)$. Note that all observations and actions are synchronized and sampled at the same sampling rate $f$.

It is well known that a single image is not sufficient as an observation to control a dynamic system such as a car \cite{dqn:mnih2015human,drive:surroundview:route:planner}, thus we use the $K$ last frames as an input to our driving model. To denote this sequence of observations we use the following notation $O_{[t-K+1,t]} := \langle O_{t-K+1}, ..., O_{t}\rangle$. Thus, we can now define our basic end-to-end driving model $\pi_{\text{bd}}$, which computes the next control action given the last $K$ observations
\begin{align}
\vspace{-1mm}
    a_{t+1} = \pi_{\text{bd}}(O_{[t-K+1,t]})\,.
\vspace{-1mm}
\end{align}

Given the Drive360 dataset, described in Section \ref{sec:dataset}, it is straightforward to use imitation learning to train our end-to-end driving model, since we have both the observations and the expert (human) actions $a^*$. To formulate the problem, we assume that our dataset has $T+1$ observation and action pairs and that our deep driving model is parametrized by $\Theta_{\text{bd}}$. Thus, the model can be learned by solving the following supervised learning problem,
\begin{align}
    \Theta_{\text{bd}}^{\star} = \arg \min_{\Theta_{\text{bd}}} \quad \sum_{t=K}^{T} \| a^*_{t+1} - a_{t+1} \| \,. \label{eq:imitation_learning}
\end{align}

\begin{figure}[tb]
\vspace{-1mm}
\centering
\includegraphics[width=0.9\linewidth]{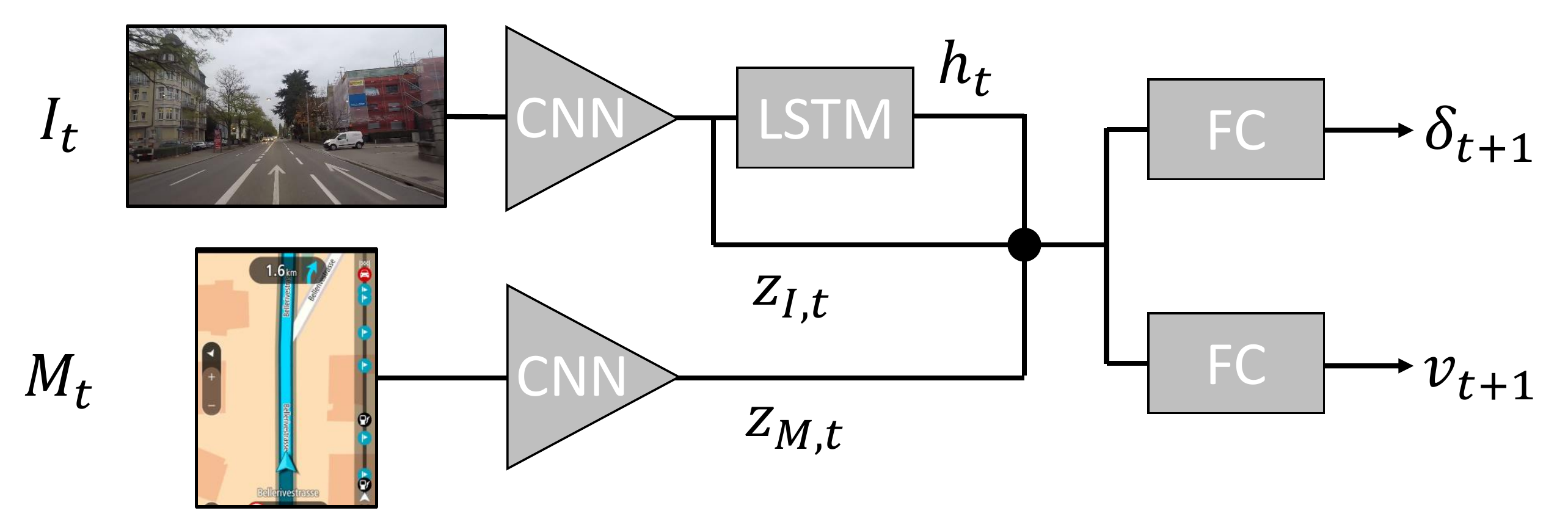}
\vspace{-4mm}
\caption{Basic end-to-end driving model.} 
\label{fig:basic_policy}
\vspace{-2mm}
\end{figure}

The architecture of the network is similar to the one used in \cite{drive:surroundview:route:planner}, without using the surround-view cameras. Thus, the image is fed through a visual encoder, and the resulting latent variable $z_{I,t}$ is further processed with an LSTM, which results in a hidden state $h_t$. The map renders are also fed through a visual encoder, resulting in a latent variable $z_{M,t}$. The three variables, $z_{I,t}$, $z_{M,t}$, and $h_t$, are then concatenated and two fully connected networks predict the actions, the full network can be seen in Figure \ref{fig:basic_policy}.

\setlength{\textfloatsep}{10pt}

\subsection{Driving Model with Richer Map Information}

As highlighted in Section \ref{sec:dataset:visual_render} we have augmented the existing Drive360 dataset to include more precise map renders that exhibit higher temporal synchronization and can replace the existing rendered navigation instructions (TomTom images). In addition, as described in Section \ref{sec:dataset:semantic_maps}, we have augmented the Drive360 dataset to include semantic map information such as distance to traffic lights, intersections, crosswalks, speed limits and road curvature etc. We now present two methods to include this semantic map information into our driving models.
%We now present two methods that improve the accuracy of a driving policy by including semantic map information. 

\subsubsection{Semantic Maps via Late Fusion}
\label{sec:approach:sem_map}
The first, albeit naive, approach is to append the semantic map information to our observations. We call this the late fusion approach as we are appending the data towards the end of the model pipeline. We format the semantic map information from all\footnote{see Section \ref{sec:experiment} for a ranking of individual group contribution.} groups as a vector and denote this vector at time step $t$ as $n_t$. This allows formulating a new observation for the driving model $O^m_t = (I_t, M_t, n_t)$. Given the new observation we can define a driving model $\pi_\text{m}(O^m_{[t-K+1,t]})$. 

\setlength{\textfloatsep}{10pt}
\begin{figure}[tb]
\centering
\includegraphics[width=0.9\linewidth]{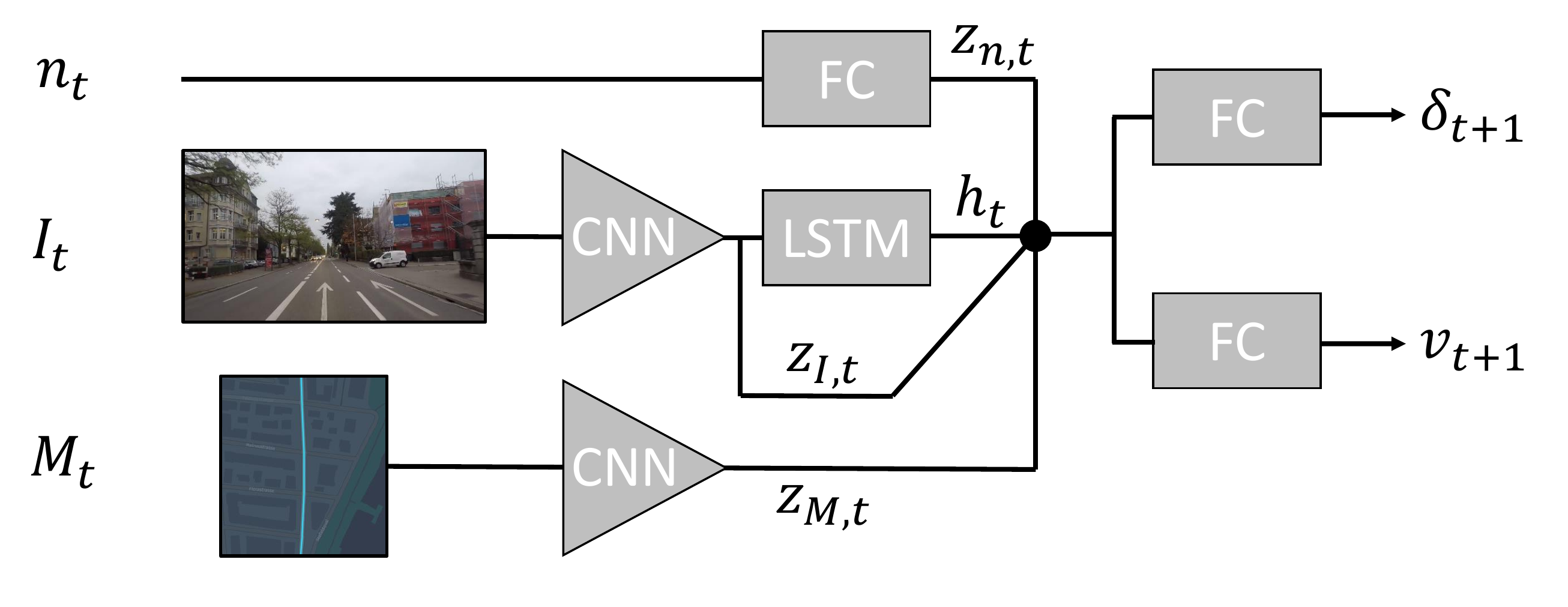}
\vspace{-2mm}
\caption{End-to-end driving model using semantic map data} 
\label{fig:semantic_policy}
\vspace{-2mm}
\end{figure}

To use the map information we introduce a new sub-module in our driving model. Therefore, $n_t$ is processed using a fully connected network and the resulting latent variable $z_{n,t}$ is concatenated with $z_{I,t}$, $z_{M,t}$, and $h_t$, before the actions are computed, see Figure \ref{fig:semantic_policy} for an illustration. 

We have investigated other network architectures such as recurrent neural networks and/or feeding in a temporal sequence of $n_t$, but did not observe further improvement. Thus, we opt for the single time-point, fully connected approach. % where the network at time-step $t$ only has access to $n_t$.

% FC semantic map network (10,10,10) or does each group have an FC?

\subsubsection{Semantic Map Attention}
\label{sec:approach:attention}
Semantic map information is very informative and useful for driving. For example, human drivers are usually alerted ahead of time, by a street sign, to an upcoming pedestrian crosswalk. This allows the human driver to pay special attention to any pedestrians in the vicinity. Thus only employing semantic maps in the late fusion stage of the network, as done in Section \ref{sec:approach:sem_map}, seems blunt. 
Ideally we would like to fuse this semantic map information during the perception stage as well, such that the driving model, similar to the human driver, is alerted to focus attention on important objects for the scenario ahead. 

We propose a method that promotes output class probabilities of a segmentation network based on semantic map information. Our approach uses a semantic segmentation network~\cite{semantic_cvpr19} pretrained on the Cityscapes dataset~\cite{Cityscapes} that generates a confidence mask for all $19$ classes. These masks are then promoted using a soft attention network that takes semantic map information, such as distance to a crosswalk or to a traffic light and the hidden state of the LSTM as input, and outputs a 19 class attention vector. Thus, the attention network allows us to boost individual class probabilities of the segmentation network based on semantic map information. The idea being, similar to how a sign might warn a human driver of an upcoming crosswalk and consequently focusing the human driver on detecting pedestrians, our attention network can boost the probability of the pedestrian class if we are close to a pedestrian crossing. The same for traffic light detection when the map tells that the car is approaching a traffic light.  

\setlength{\textfloatsep}{10pt}
\begin{figure}[tb]
\vspace{-2mm}
\centering
\includegraphics[width=0.9\linewidth]{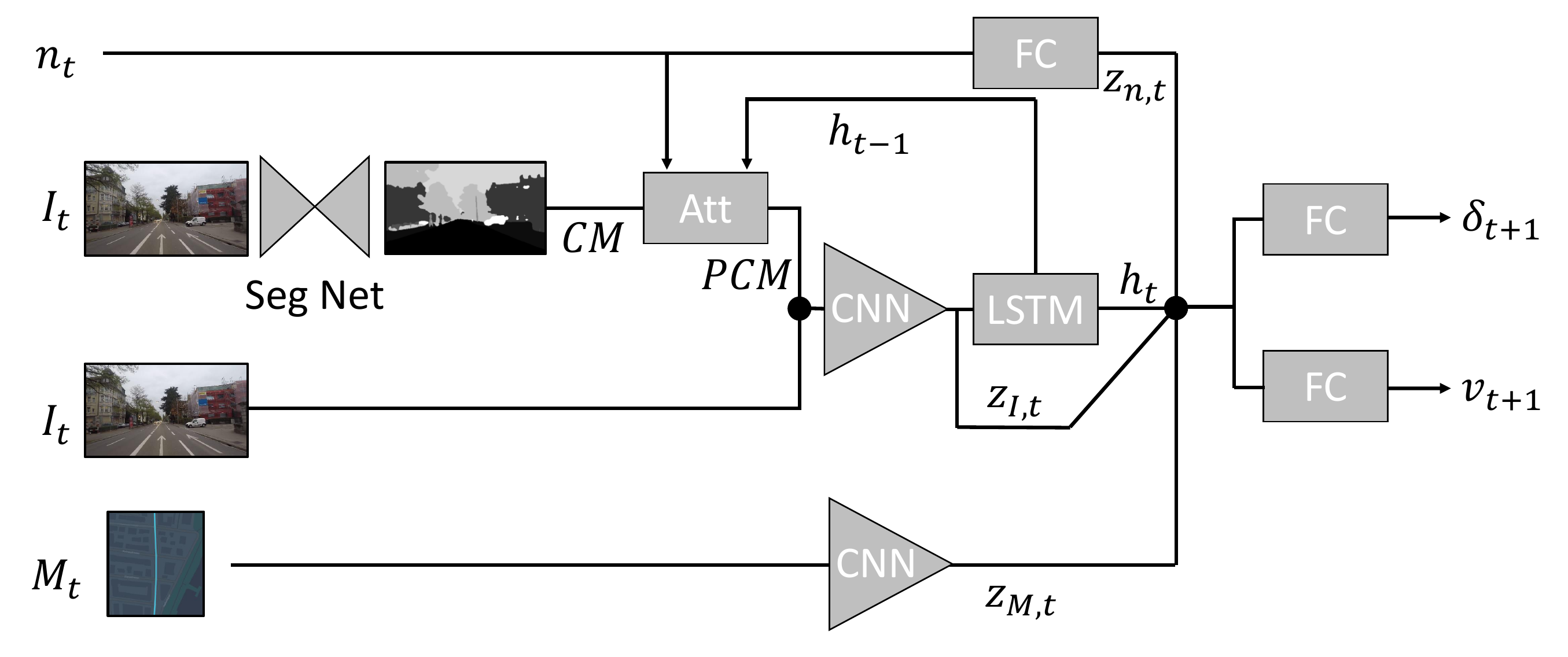}
\vspace{-4mm}
\caption{End-to-end driving model using semantic map data and attention} 
\label{fig:attention_policy}
\vspace{-2mm}
\end{figure}

More precisely our complete architecture is shown in Figure \ref{fig:attention_policy}. Compared to the network used in the previous section, the main difference is the addition of the segmentation network and the attention network. This part first passes the front-facing camera image through the segmentation network, which generates confidence masks for all classes. The output of the network are 19 confidence masks, for classes such as pedestrians, road, and traffic signs. We denote the $i$-th confidence mask as $CM_i$. A direct use of the segmentation results may not be optimal as the method might misclassify the important objects for the current situation or completely misses them due to low confidence score.  
%As such, the segmentation masks do not give significant additional information to the driving network since the information is already encoded in the input image. 
Therefore, we use an attention network to promote the different confidence masks, where the promotion is based on the numeric map information and the previous hidden state of the LSTM. The promoted confidence mask of the $i$-th class $PCM_i$, is computed as follows:
\begin{align}
    &\alpha = f(n_t,h_{t-1})\,, \nonumber\\ 
    &a = \frac{\alpha}{\max(\|\alpha\|_2,\epsilon)}\,, \nonumber\\
    &PCM_i = a_i \cdot CM_i\,, \label{eq:attention}
\end{align}
where, $\alpha \in \mathbb{R}^{19}$ is the attention vector and $a \in \mathbb{R}^{19}$ normalized promotion vector, $f$ is a fully connected network, and $\epsilon = 10^{-19}$ is added to avoid divisions by zero. Note that we use a normalization, compared to the more popular soft max to generate $a$, as it is likely that several classes are important at each time step.

The rest of the network architecture stays the same compared to the model with semantic maps, and we denote the resulting driving model as $\pi_\text{am}(O^m_{[t-K+1,t]})$. Note that both semantic map based networks can be trained using imitation learning, as formulated in \eqref{eq:imitation_learning} for the basic driving model.
\subsection{Human-Like Driving}
\label{sec:approach:human_like}
When training the driving model as presented in \eqref{eq:imitation_learning}, the decision problem is treated as a supervised regression problem with i.i.d. targets (expert actions). However, in reality, this is not the case and actions taken in the current time instant influence the actions in the next time steps. Including this temporal dependency is fundamental while training an end-to-end driving model, thus, we propose matching action sequences (which we call drivelets) instead of single actions. We denote a drivelet for the next $N$ time steps by $\mathbf{a}_{t} = \langle a_t^T,...,a_{t+N}^T \rangle$, where $a_t$ is computed using one of the previously proposed driving models. Given a drivelet, the imitation learning problem can be posed as,
\begin{align}
    \Theta^\star = &\arg \min_{\Theta} \quad \sum_{t=K}^{T-N} \| \mathbf{a}^*_{t+1} - \mathbf{a}_{t+1} \| \,.
\end{align}
However, simply matching a sequence of actions does not change the regression problem, since there is no loss on the temporal information of the drivelet. In this section we propose to formulate the problem as a Generative Adversarial Network (GAN)~\cite{gan:goodfellow}, where the driving model is the generator, and the discriminator judges if a drivelet is from the same distribution as human drivelets. %\simoncomment{Could be removed as already part of the intro} This has two fundamental advantages, first, the GAN now acts on the whole drivelet, adding a loss on the temporal information of the drivelet. Secondly, and arguably more important, the discriminator forces the driving model, to generate action sequences that are human-like. A driving model that acts like a human should result in a driving style passengers consider comfortable. Furthermore, a car that behaves human-like is preferable in traffic with mixed human and self-driven cars, as the behavior is easier to predict by human drivers, which in terms should simplify the acceptance of autonomous cars.

To formulate the GAN problem, we introduce a discriminator $\Delta$, consisting of a fully connected network, parameterized by $\nu$. The discriminator takes a drivelet $\mathbf{a}_{t}$ as input, and outputs the probability of classifying this drivelet as human driving. Using the standard GAN learning framework, the discriminator tries to correctly classify human and machine drivelets, whereas the driving model tries to fool the discriminator into thinking its drivelets are generated by a human driver. Thus, we define our loss for human-like driving as
\begin{equation}
    \label{eq:objective:adverse} 
    \mathcal{L}_t^{hum} = - \text{log}(\Delta(\mathbf{a}_{t}))\,.
\end{equation}
Combining the human-like GAN loss, with the imitation loss, we can formulate our new human-like imitation learning problem, as a zero-sum game between the driving policy and the discriminator, 
\begin{align}
    \min_{\Theta} \: \max_{\nu}  \sum_{t=k}^{T-N} \| \mathbf{a}^*_{t} - \mathbf{a}_{t]} \| + \lambda \mathcal{L}_t^{hum}\,,
\end{align}
where $\lambda$ allows to trade-off imitation of the human driver and generating human-like drivelets. 

\section{Experiments}
\label{sec:experiment} 

%The aim of this section is to demonstrate the benefit of the components proposed in this work which can be added to a standard end-to-end driving model. 

\subsection{Implementation Details}
Our driving model consists of multiple submodules:

\noindent 
\textbf{(1) Front-Camera Module}: Our core image module consists of a Resnet18~\cite{resnet} CNN to process sequences of front facing camera images. The resulting 2096 element latent variable $z_{I,t}$ is fed to an LSTM with 64 hidden states. We opt for a Resnet18 architecture due to its small size, whereas larger CNNs can be substituted for higher accuracy.

\noindent 
\textbf{(2) Front-Camera Attention Module}: We use the segmentation network proposed by ~\cite{semantic_cvpr19} pre-trained on Cityscapes to generate a $19$ class confidence mask by removing the argmax layer. This confidence mask is channel-appended with the front facing camera image resulting in a $22$-channel ``image" that is processed in a Resnet18~\cite{resnet} CNN where the first layer is modified to accept $22$ channels instead of $3$. The fully connected network in the attention mechanism \eqref{eq:attention}, is a three layer network with 128 - 64 - 19 neurons.

\noindent 
\textbf{(3) Visual Map Module}: Following~\cite{drive:surroundview:route:planner}, a fine-tuned AlexNet \cite{alexnet} is used to process the rendered navigation maps, either from TomTom or HERE. 

\noindent 
\textbf{(4) Semantic Map Module}: A fully connected network with three layers and 30 neurons per layer is used to to process all semantic map information. 

\noindent
\textbf{(5) Control Module}: Following~\cite{drive:surroundview:route:planner}, the control module consists of two action heads that predict either steering angle or speed. Each action head is composed of a fully connected network with three layers, and 64 - 64 - 32 neurons.

\noindent 
\textbf{(6) Human-like Module}: A fully-connected, three-layer discriminator network with 10 neurons per layer is used to classify the human-like driving.

Combinations of these sub-modules allow us to define the three driving model architectures proposed, along with any intermediate driving model representations, see Table \ref{table:evaluation}. As a note, all models include the control module.

In Table \ref{table:evaluation}, $Model_4$ is our late fusion semantic map model, corresponding to the model proposed in Section~\ref{sec:approach:sem_map}. It uses the front-facing camera, visual map and semantic map module. As a build up to $Model_4$ we define $Models_{1-3}$ to validate the late fusion approach. $Model_6$ is our attention network approach, corresponding to Section~\ref{sec:approach:attention}. This model incorporates the front-camera attention model along with the semantic and visual map modules for guidance. As validation to $Model_6$ we propose $Model_5$ which is similar but includes the segmentation masks without the attention module. $Model_7$ is our adversarial human-like model, corresponding to Section~\ref{sec:approach:human_like}. It incorporates the front camera, visual and semantic maps, and the human-like module. Finally we present $Model_8$ which is combination of all sub-modules. 

Each driving model is trained on a subset (580k sequences) of the Drive360 + HERE dataset. The subset is selected using the semantic map annotation and includes a more even distribution of curved to straight roads, as well as traffic light and pedestrian crossing situations. This subset speeds up the training process by a factor of 3 compared to the full dataset while not significantly sacrificing accuracy. We train with a batch size of 8 for three epochs on an Nvidia V100 GPU. 
%Training for more epochs does not significantly improve convergence.

\begin{table*}[]
\vspace{1mm}
\footnotesize 
\setlength\tabcolsep{1.2pt}
\begin{tabular*}{\textwidth}{lcccccc @{\extracolsep{\fill}}ccccccccc} \toprule
 \multirow{3}{*}{\rot{Model id}} &
 \multirow{3}{*}{\rot{Camera}} & \multirow{3}{*}{\rot{Render}} & \multirow{3}{*}{\rot{SemMap}} & \multirow{3}{*}{\rot{SegMask}} & \multirow{3}{*}{\rot{Attention}} & \multirow{3}{*}{\rot{Hu-like}}  & \multicolumn{3}{c}{$\mathbb{S}$: the whole } & \multicolumn{2}{|c}{$\mathbb{A}$: traffic light} & \multicolumn{2}{|c}{$\mathbb{B}$: curved } & \multicolumn{2}{|c}{$\mathbb{C}$: approaching} \\
  \multicolumn{7}{r}{}  & \multicolumn{3}{c}{ evaluation set} 
 & \multicolumn{2}{|c}{or pedestrian line } 
 & \multicolumn{2}{|c}{mountain road} 
 & \multicolumn{2}{|c}{intersections}\\
 \multicolumn{7}{r}{} & $A_\delta$ $\downarrow$ & $A_v$ $\downarrow$ & $H$ $\uparrow$ & $A_\delta$ $\downarrow$ & $A_v$ $\downarrow$ & $A_\delta$ $\downarrow$ & $A_v$ $\downarrow$ & $A_\delta$ $\downarrow$ & $A_v$ $\downarrow$ \\ \midrule

\cite{end:driving:16} & \YES & & & & &   & 1273 & 8.37  & 53.8  & 1100 & 10.65  & 4351 & 3.00  & 3322 & 8.66 \\
\cite{drive:surroundview:route:planner} & \YES & T & & & &   & 1222 & 5.89  & 56.7  & 689 & 7.20  & 3198 & 2.48  & 2579 & 6.34 \\
1 & \YES & & & & &   & 1419 & 7.24  & 55.7  & 1121 & 8.46  & 4132 & 2.93  & 3432 & 8.05 \\
2 & \YES & H & & & &   & 1166 & 6.63  & 57.1  & 710 & 8.74  & 1334 & 2.37  & 2128 & 7.65 \\
3 & & H & \YES & & &   & 1094 & 18.44  & 47.7  & 969 & 14.05  & 2680 & 3.45  & 2797 & 14.50 \\
4 & \YES & H & \YES & & &   & 1035 & 5.25  & 57.4  & 781 & 5.05  & 2562 & 4.57  & 2397 & 5.62 \\
5 & \YES & H & \YES & \YES & &   & 997 & 5.56  & 55.8  & 585 & 3.44  & 1979 & 3.25  & 1928 & 5.74 \\
6 & \YES & H & \YES & \YES & \YES &   & 940 & \textbf{4.97}  & 57.7  & 598 & \textbf{3.11}  & 1634 & 2.32  & 2021 & \textbf{5.28} \\
7 & \YES & H & \YES & & & \YES   & \textbf{932} & 5.53  & \textbf{60.7}  & \textbf{520} & 5.82  & \textbf{1211} & 2.65  & \textbf{1898} & 6.15 \\
8 & \YES & H & \YES & \YES & \YES & \YES   & 980 & 5.03  & 58.0  & 693 & 3.33  & 1705 & \textbf{2.29}  & 2111 & 5.42 \\
\bottomrule
\end{tabular*}
\caption{The performance of all variants of our method evaluated on the four evaluation sets defined. Driving accuracy (MSE error) is denoted by $A_\delta$ \& $A_v$ for steering angle (degree) and speed (km/h) respectively, and the human-likeliness score by $H$ (\%). $\uparrow$ means that higher is better and $\downarrow$ the opposite.}
\label{table:evaluation}
\vspace{-10mm}
\end{table*}

\subsection{Evaluation}
A driving model should drive as accurately as possible in a wide range of scenarios. As our models are trained via imitation learning, we define accuracy as how close the model predictions are to the human ground truth maneuver under the L2 distance metric. We define $A_\delta$ as the mean squared error in the steering angle prediction and $A_v$ as the mean squared error in the vehicle speed prediction. Specifically, we predict the steering wheel angle $\delta_{t+1s}$ and vehicle speed $v_{t+1s}$ 1s into the future \footnote{Predicting further into the future is possible and our experiments have shown a growing degradation in accuracy the further one predicts.}. We use a SmoothL1 loss to jointly train $\delta_{t+1s}$ and $v_{t+1s}$ using the Adam Optimizer with an initial learning rate of $10^{-4}$.

\textbf{Evaluation Sets}. 
We denote the complete Drive360 test set by $\mathbb{S}$, consisting of around 10 hours of driving, covering a wide range of situations including city and countryside driving. While one overall metric, such as MSE, on the whole test set is easier to interpret, evaluations on specific scenarios can highlight the strengths and weaknesses of driving models at a finer granularity.
%\TODO{We observe that our GAN model... is the best! Give some elaborate explanation as to why.
% \footnote{Appenzell\_3, Bern, Luzern} 
By enriching the Drive360 dataset with HERE map data, we can filter the test set $\mathbb{S}$ for specific scenarios. We have chosen three interesting scenarios in this evaluation:
\vspace{-1mm}
\begin{itemize}
\itemsep -0.5\parsep
\item $\mathbb{A} \subset \mathbb{S}$ where the distance to the next traffic light is less than 40m or the distance to the next pedestrian crossing is less than 40m and the speed limit is less than or equal to 50km/h. Translates to \textit{approaching a traffic light or pedestrian crossing in the city}.
\item $\mathbb{B} \subset \mathbb{S}$ where the curvature is greater than 0.01 and the speed limit is 80 km/h and the distance to the next intersection greater than 100m. Translates to \textit{winding road where road radius is less than 100m and no intersections in the vicinity}.
\item $\mathbb{C} \subset \mathbb{S}$ where the distance to the next intersection is less than 20m, named \textit{approaching an intersection}. 
\end{itemize}
\vspace{-1mm}
%\TODO{We observe that... Reasoning being...}

\textbf{Comparison to state-of-the-art}. 
We compare our method to two state-of-the-art end-to-end driving methods~\cite{end:driving:16} and \cite{drive:surroundview:route:planner}. They are trained under the same settings as our method is trained. The results are shown at the top of Table~\ref{table:evaluation}. Our three approaches achieve significant performance gains over the two competing methods. The following sections will now evaluate each approache in detail.   

\subsubsection{Semantic Maps}
We have presented two methods that make use of the semantic maps, late fusion and attention based.
\noindent
\textbf{Late-Fusion.} By comparing the performance of $Model_{1-4}$ in Table~\ref{table:evaluation}, one can find that driving accuracy improves significantly when using maps in general and semantic maps in particular. The best results are achieved by using the combination of semantic and visual maps as done in $Model_4$.
The benefit of semantic and visual maps are most dramatically illustrated in $Model_3$, which does not incorporate front-facing camera information at all to predict driving maneuvers. $Model_3$ performs competitively for steering angle prediction, yet fails dramatically for speed prediction.
This is in line with the intuition that, on average, mapping road curvature to steering commands leads to accurate driving, because we usually do not perform evasive maneuvers. It goes without saying that $Model_3$ would fail horribly if an evasive maneuver is required or vehicle localization fails. As speed is extremely dependent on other vehicles in the vicinity we observe the drop in speed prediction performance due to lack of camera observations. 
%Nonetheless this toy example not only demonstrates the effectiveness of semantic maps for steering angle prediction but could also aid in the search for rare driving scenarios as these would 

The best results are achieved by using semantic and visual maps together.
% The TomTom component, used exclusively, offers general accuracy improvements over HERE features, used exclusively. However in specific scenarios HERE features are superior, see $A_s$ at $\mathbb{C}$.
We reason that the visual map module offers a complete world view, in other words, an aggregation of all road geometries. It is designed to facilitate human driving, but it seems that neural networks benefit from having this intuitive representation as well.

The designed features out of our semantic map data are in stark contrast to the rendered visual representation. They are accurate and unequivocal. By using semantic map features, our method $Model_4$ outperforms \cite{drive:surroundview:route:planner} which only uses rendered visual maps. It also outperforms \cite{end:driving:16} which uses no map information. The used semantic map features are all relevant to driving. Individual feature groups, however, contribute to a different extent. We experimentally verified individual feature group contribution and have the following ranking for speed prediction $2{>}1{>}6{>}5{>}3{>}4$ and steering angle prediction $3{>}6{>}1{>}2{>}4{>}5$, in line with intuition. 

\textbf{Semantic Map Based Attention.}
$Model_5$ adds segmentation network probability masks and improves upon $Model_4$ in terms of steering and speed prediction. Thus segmentation masks on their own are beneficial. However, we would like to validate whether our semantic map attention can boost driving model performance further. Indeed, $Model_6$, which adds in our attention mechanism, further improves over $Model_5$, showing that it is not only the presence of the segmentation mask that is responsible for the performance gains but indeed the attention mechanism. 

\setlength{\textfloatsep}{10pt}
\begin{figure}[ht]
\centering
\includegraphics[width=0.90\linewidth]{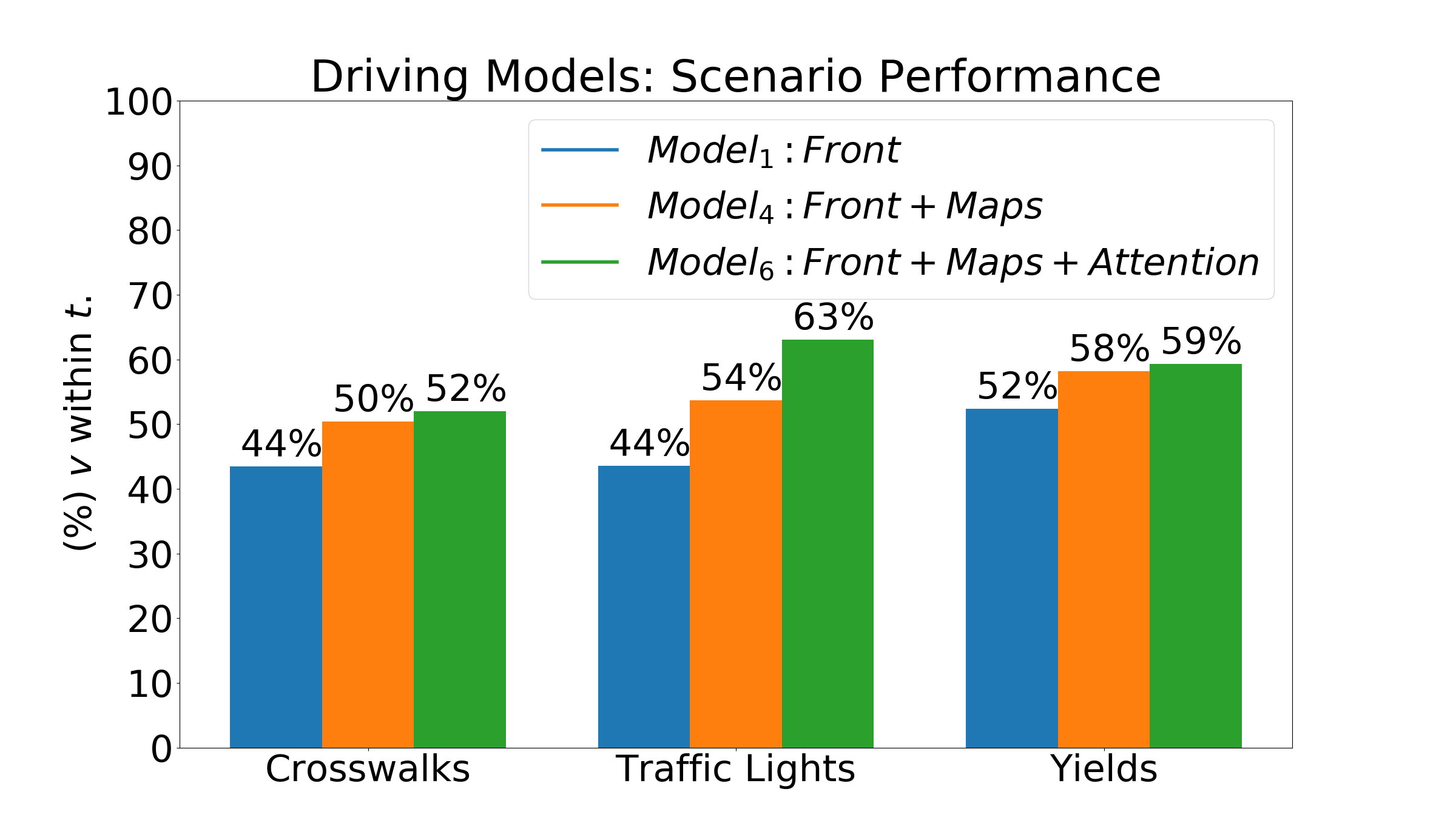}
\vspace{-6mm}
\caption{Percentage of correct speed ($v$) predictions using a $1m/s$ tolerance band ($t$).} 
\label{fig:evaluation:traffic_light}
\vspace{-3mm}
\end{figure}

Utilizing the semantic maps for subset selection, we dive into a more fine grained evaluation illustrated in Figure \ref{fig:evaluation:traffic_light}. Here we compare the performance of $Model_1$ (front-camera only), $Model_4$ (visual+semantic maps) and $Model_6$ (attention model) in three unique driving situations. These driving situations are defined as the vehicle being within 80m of an upcoming crosswalk, a traffic light or a yield instance. We chose these situations as we believe in these situations the benefit of the semantic map attention module will become even more apparent. For easier human understanding we opt for a tolerance band evaluation, defining correct model speed predictions if they are within 1m/s of the human ground truth.
Indeed for all instances $Model_6$ outperforms the two others. The most notable scenario being traffic lights. $Model_5$, utilizing semantic maps, already results in an absolute 10\% performance boost, however with our attention model, we are able to boost absolute speed prediction performance by almost 20\%. These are significant gains and highlight how the map based attention model can overcome segmentation network shortcomings by promoting classes based on map-defined driving situations. 
The intuition being that the class probability of the traffic light, in this example, will be promoted when the semantic map indicates the presence of one in near future. This, to some extent, relaxes the performance constraint on segmentation networks for autonomous driving.
As opposed to existing attention mechanisms for driving models~\cite{attention:objectcentric:ad:icra19} that require accurate object detection, our presented method can promote the relevant class even if there is only a minuscule detection probability.  

\subsubsection{Human-likeness}
When employing the human-like training, our discriminator network is tasked with classifying maneuvers either as being human or machine created using a cross entropy loss. 
It is hard to quantify weather a driving style is human-like. It is also hard to evaluate. In order to evaluate it quantitatively, we propose a new evaluation criterion -- the \textit{human-likeness score}. This score is calculated by clustering human driving maneuvers ($\delta$ and $v$ concatenated) from the evaluation set $\mathbb{S}$, over a $2$s window with a step size of $1$s, into 50 unique clusters using the Kmeans algorithm. Predicted model maneuvers are then considered human-like if, for the same window, they are associated with the same cluster as the human maneuver. We chose our window and step size to be consistent with our model training. The \textit{human-likeness score} is then defined as the percentage of driving maneuvers generated by a driving model that are associated to the same cluster as the human driving maneuvers for the same time window. 

To this end we generate model driving maneuvers via a sequence of three ($N=3$) consecutive steering wheel angle and vehicle speed predictions.
% To promote human-like driving, we task our driving model to generate maneuvers that the discriminator is unable to differentiate from human maneuvers. This is achieved by calculating the inverse loss of a model generated maneuver given a human maneuver label. This adversary loss is added to the existing loss.
We observe, again in Table~\ref{table:evaluation}, that our adversarial learning designed for learning a human-like driving style improves performance and in particular boosts steering accuracy and the human-likeness score, see $Model_{7}$ vs $Model_{4}$. 
%For example, the human-likeness score notably increases between $Model_{1}$ vs $Model_{7}$ resulting in additional driving accuracy gains. 
Interestingly, when a model drives more accurately, due to the presence of a navigation component, its human-likeness score improves as well. This is evidenced by the performance comparison of $Model_1$ vs. $Model_{4}$ and \cite{end:driving:16} vs. \cite{drive:surroundview:route:planner}. We believe that this due to the driving models clearer understanding of the driving environment which consequently yields to quite human-like driving already. 
% However, we observe similar gains in driving accuracy when supplying either the navigation components as in $Model_{4}$ or the human-likeness component as in $Model_{7}$ to our baseline $Model_{1}$.
Overall, the model trained using all our modules performs comparably to $Model_{6}$ and $Model_{7}$, and we believe that this model has the most promise for future work as it can combine the strengths of both parent models.

\section{Conclusion}
This work demonstrates the usefulness of semantic maps for end-to-end driving models and an adversarial training strategy to promote human-like driving. We have presented two approaches that incorporate semantic maps which we provide for the Drive360 dataset: 1) our late fusion approach boosts model performance over the standard top-down rendered navigation approach and 2) our semantic map attention mechanism further improves model performance, with large gains observed in specific driving scenarios such as at traffic lights. Additionally, semantic maps allow us to filter for specific driving scenarios, offering finer evaluation control and leading to greater model interpret-ability.
Adversary learning was introduced such that the learned driving model behaves more like human drivers. Our proposed models are more accurate and human-like than previous methods. 

{\small
\bibliographystyle{IEEEtran}
\bibliography{egbib}
}
\end{document}